%% file: main_preprint.tex
\documentclass[conference]{IEEEtran}
\IEEEoverridecommandlockouts

\usepackage{cite}
\usepackage{amsmath,amssymb,amsfonts}
\usepackage{algorithmic}
\usepackage{graphicx}
\usepackage{textcomp}
\usepackage{xcolor}
\usepackage{soul}
\def\BibTeX{{\rm B\kern-.05em{\sc i\kern-.025em b}\kern-.08em
    T\kern-.1667em\lower.7ex\hbox{E}\kern-.125emX}}

\usepackage[
placement=top,
angle=0,
color=red,
scale=1.1,
hshift=0,
vshift=-15
]{background}
\backgroundsetup{contents=\bf{----------------------------------------------------------------- PREPRINT -----------------------------------------------------------------}}

\begin{document}

\def\NoNumber#1{{\def\alglinenumber##1{}\State #1}\addtocounter{ALG@line}{-1}}
\definecolor{mygreen}{RGB}{0,204,102}
\newcommand{\ACe}[1]{\textcolor{red}{#1}}
\newcommand{\rev}[1]{\textcolor{blue}{#1}}
\title{\vspace*{18pt}Smart Fleet Solutions: Simulating Electric AGV Performance in Industrial Settings}
\author{
    \IEEEauthorblockN{Tommaso Martone\IEEEauthorrefmark{1}, Pietro Iob\IEEEauthorrefmark{1}\IEEEauthorrefmark{2}, Mauro Schiavo\IEEEauthorrefmark{2}, Angelo Cenedese\IEEEauthorrefmark{1}}
    \IEEEauthorblockA{\IEEEauthorrefmark{1}\textit{Department of Information Engineering, University of Padova, Padua, Italy}
    \\\ tommaso.martone@studenti.unipd.it, pietro.iob@phd.unipd.it, angelo.cenedese@unipd.it}
    \IEEEauthorblockA{\IEEEauthorrefmark{2} \textit{Techmo Car S.p.a., Padua, Italy}
    \\\ pietro.iob@techmo.it, mauro.schiavo@techmo.it}
}
\maketitle
\begin{abstract}


This paper explores the potential benefits and challenges of integrating Electric Vehicles (EVs) and Autonomous Ground Vehicles (AGVs) in industrial settings to improve sustainability and operational efficiency. While EVs offer environmental advantages, barriers like high costs and limited range hinder their widespread use. Similarly, AGVs, despite their autonomous capabilities, face challenges in technology integration and reliability. To address these issues, the paper develops a fleet management tool tailored for coordinating electric AGVs in industrial environments. The study focuses on simulating electric AGV performance in a primary aluminum plant to provide insights into their effectiveness and offer recommendations for optimizing fleet performance. 

\end{abstract}
\IEEEpeerreviewmaketitle

\input{chapters/01_Introduction}      
\input{chapters/02_ProblemDescription}
\input{chapters/03_ProposedSolution}
\input{chapters/04_SimulationSetup}
\input{chapters/05_Results}
\input{chapters/06_Conclusions}

\section*{Acknowledgment}
This research is supported by the collaboration of Techmo Car S.p.a. (Padova, Italy) with the University of Padova, Italy.
\bibliography{bibliography.bib} 
\bibliographystyle{ieeetr}
\end{document}

%% file: chapters/01_Introduction.tex
\section{Introduction}
\label{sec:Introduction}

The surge in Electric Vehicles (EVs) deployment reflects a shift towards sustainable transport, particularly noticeable in industrial contexts \cite{buekers2014health}. However, challenges like battery technology limitations hinder widespread adoption \cite{deng2020electric}. EVs offer environmental benefits but face hurdles like high costs and limited range in industrial use. Autonomous Ground Vehicles (AGVs) offer logistical solutions, but also face adoption challenges akin to EVs \cite{stefanini2022environmental,javed2021safe}. This paper aims to tackle these issues by developing a plant simulator tool for AGV coordination. It aims to showcase electric AGVs' capabilities in industrial settings and serve as a performance evaluation tool for fleet deployment \cite{bielli2011trends}.

Specifically, without sacrificing applicability, this study targets the creation of a simulation environment capable of assessing the performance of such a fleet within the confines of a primary aluminum plant. Section \ref{sec:ProblemDescription} will explore a comprehensive elucidation of the problem domain and the author's proposed solution aimed at materializing the envisioned simulation environment, while a specific emphasis on the selected case study will be presented in \ref{sec:SimulationSetup}. Finally, Section \ref{sec:Results} will encapsulate the principal findings gleaned from the ensuing simulation campaign, thereby encapsulating the essence of this endeavor.

%% file: chapters/02_ProblemDescription.tex
\section{Problem Description and Proposed Solution}
\label{sec:ProblemDescription}
This work can be used as a tool to assess multiple performance levels. However, as a valuable example, this paper will focus on determining the minimum size a fleet must have to fulfill all plant requests and prevent accumulation. Determining the optimal fleet size is influenced by plant priorities and operational efficiency, with vehicle routing significantly impacting task completion times and overall efficiency.
The study introduces the \textit{Fleet Management Simulator} (FMS), a versatile and modular tool adaptable to various contexts, providing insights into the operational dynamics of electric AGVs-driven plants.
The FMS is structured as a \textit{Finite State Machine} (FSM) \cite{lynch1996distributed}, with three main states: \textit{Idle}, \textit{Charging}, and \textit{Routine}. The \textit{Routine} state includes essential vehicle tasks for plant functioning, the \textit{Charging} state accounts for recharging needs, and the \textit{Idle} state designates availability for secondary tasks.
Task allocation is managed by the \textit{Plant Manager} (PM), which selects optimal vehicles for plant requests, and a \textit{Decentralized Tasks Manager} (DTM), which assigns temporary tasks to available vehicles when there are no plant requests. This dual management approach ensures comprehensive consideration of plant and vehicle needs.

%% file: chapters/03_ProposedSolution.tex

This work focuses on three vehicle types: \textit{Fluoride Feeder Vehicle} (\textit{FFV}), \textit{Anode Pallet Transport Vehicle} (\textit{APTV}), and \textit{Metal Transport Vehicle} (\textit{MTV}). These vehicles are essential in aluminum smelting: the \textit{FFV} distributes Aluminum Fluoride ($AlF_3$), the \textit{APTV} transports carbon anodes, and the \textit{MTV} moves molten aluminum to the furnaces.

\begin{figure}[ht]
    \centering
    \includegraphics[width=1\linewidth]{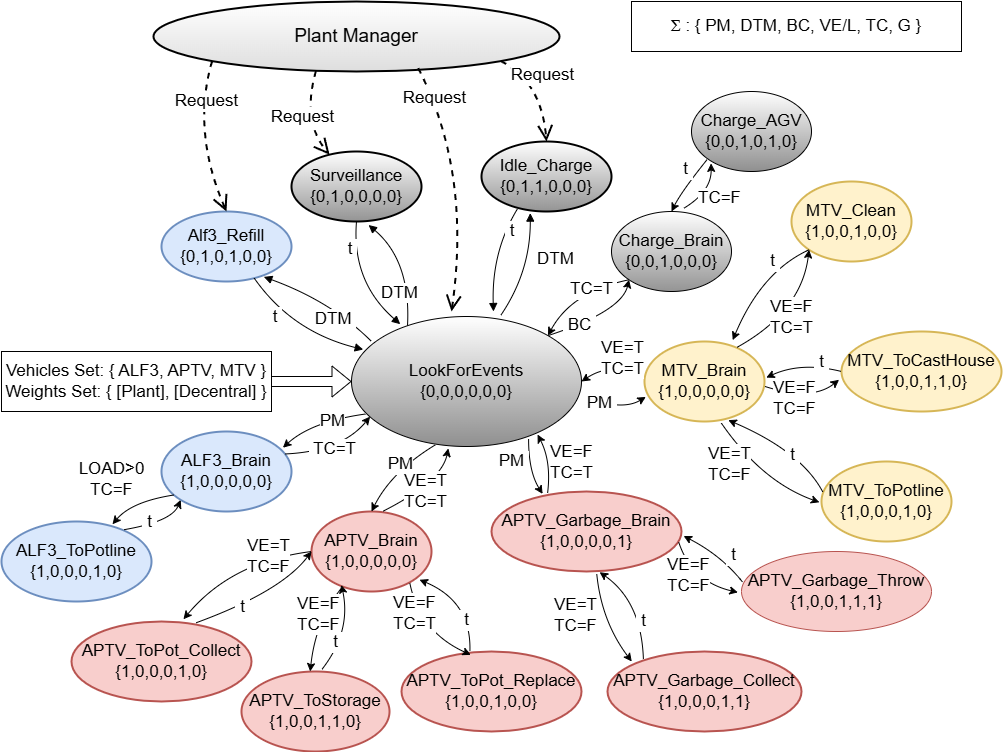}
    \caption{Finite State Machine Diagram.}
    \label{fig:LL_diagram}
\end{figure}

Fig. \ref{fig:LL_diagram} illustrates the FSM implementation, showing system inputs, vehicle quantities, and weights for the \textit{Plant Manager} and \textit{Decentralized Tasks Manager}. Colors indicate different vehicle classes: blue for \textit{FFV}, red for \textit{APTV}, and yellow for \textit{MTV}, with grey states accessible by any vehicle. \textit{Idle} states represent temporary tasks while vehicles are available for the PM. PM-reachable states are marked with a dotted line.

The transition between states is determined by a symbol alphabet defined as follows: $$\Sigma: \{PM, DTM, BC, VE/L, TC, G\}$$

\begin{itemize}
    \item PM: \textit{Plant Manager} request.
    \item DTM: \textit{Decentralized Task Manager} selection.
    \item BC: Battery Charge under 20\%.
    \item VE/L: The Vehicle is Empty or Load for \textit{FFVs}.
    \item TC: The task has been completed. 
    \item G: Specific symbol for a \textit{PM} task called \textit{Garbage task}.
\end{itemize}

The combination of these symbols with the vehicle's class will define the different state transitions of the AGVs. As previously mentioned, the three primary states can be further explored, unveiling the entire system.


In \textit{Idle} states, vehicles perform temporary tasks while available for selection by the PM. The \textit{LookForEvents} state allows vehicles to determine their next state based on battery charge and commands from the PM and DTM. In \textit{Surveillance}, vehicles scout less visited areas to monitor and gather data using stereo cameras or Lidar \cite{zhao2020fusion} \cite{cortes2004coverage}. In \textit{Idle\_charging}, vehicles recharge briefly at a charging station. The \textit{$AlF_3$\_refill} state sends \textit{FFVs} to refill at the $AlF_3$ storage.



The \textit{Charge} states include $Charge\_Brain$ and $Charge\_AGV$. Vehicles below $20\%$ SOC select a charging station in $Charge\_Brain$ and charge fully in $Charge\_AGV$. The \textit{Routine} states differ by vehicle class, involving specific tasks such as pot refilling for \textit{FFVs}, anode replacement and waste removal for \textit{APTVs}, and molten aluminum collection for \textit{MTVs}.

For the design of the \textit{Plant Manager}, a cost function model has been employed. This approach offers significant flexibility; by adjusting the weights it becomes possible to enhance the relevance of one behavior over another.
For \textit{APTV} and \textit{MTV} the employed cost function is reported in eq.\eqref{eq:APTV_MTV_fc}:
\begin{equation}
\label{eq:APTV_MTV_fc}
    f_{V} = W_{r} \cdot (R - d_{req}) + \frac{W_{d}}{d_{task}}
\end{equation}

For \textit{FFV} the employed cost function is reported in eq.\eqref{eq:FFV_fc}:
\begin{equation}
\label{eq:FFV_fc}
    f_{FFV} = f_{V} + W_{l} \cdot m_{l} 
\end{equation}

Where $R$ [m] is the estimated distance range the vehicle can cover with the current charge. $d_{req}$ [m] is the total distance required to perform the task and navigate to the closest charging station. $d_{task}$ [m] is the exact distance from the vehicle's current position to the assigned destination. $m_{l}$ [kg] is the current mass of $AlF_3$ on the \textit{FFV}. $W_{r}$, $W_{d}$, and $W_{l}$ represent the weights associated with vehicle autonomy, goal distance, and vehicle load, respectively. 
The estimation of $R$ is conducted through the following computations:
\begin{equation}
    \label{eq:Autonomy1}
    R = SOC \times d_{avg} \qquad \qquad d_{avg} = \frac{d_{tot}}{E_{c}- E_{r}}  \\
\end{equation}

Where $d_{avg}$ [m/Wh] is the average distance a vehicle can cover per unit of energy, while $d_{tot}$ [m] is the total distance covered and $E_{c}$ [Wh] is the total energy consumed. Finally, $E_{r}$ [Wh] is the total energy regenerated through regenerative brakes and $SOC$ [Wh] is the state of charge of the vehicle.
The accuracy of $d_{avg}$ improves throughout the simulation as parameters in eq.\eqref{eq:Autonomy1} are calculated at each step, increasing accuracy with more data.
The \textit{DTM} determines the best task for each vehicle based on their charge, position, and for the \textit{FFV}, the amount of $AlF_3$ in its tank, unlike the \textit{PM}, which compares all vehicles.
Each vehicle uses a decentralized cost function tailored to specific tasks. By comparing these tailored functions, each vehicle identifies the task with the minimum cost. 

The employed decentralized cost functions are reported in the following eq.\eqref{eq:decentralized1}, \eqref{eq:decentralized2}, \eqref{eq:decentralized3}:
\begin{align}
    \label{eq:decentralized1}
    &f_{charge} = W_{SOC} \cdot SOC + W_{dist} \cdot (d_{CS} + V_{path}) \\
    \label{eq:decentralized2}
    &f_{surv} = W_{surv} \cdot V_{min} + W_{dist} \cdot (d_{E} + V_{path}) \\
    \label{eq:decentralized3}
    &f_{refill} = W_{Load} \cdot m_{l} + W_{dist} \cdot (d_{R} + V_{path})
\end{align}

where $V_{path}$ is the sum of visit values, $d_{CS}$ is the distance to the charging station, $d_{E}$ is the distance to the least visited area. Moreover, $d_{R}$ [m] is the distance to the closest $AlF_3$ storage and $V_{min}$ is the minimum visit value. Finally, $W_{SOC}$, $W_{d}$, $W_{surv}$, and $W_{Load}$ are weights for vehicle SOC, goal distance, plant surveillance, and vehicle load.

%% file: chapters/04_SimulationSetup.tex
\section{Simulation Setup}
\label{sec:SimulationSetup}
The simulations utilized the open-source software SUMO (Simulation of Urban Mobility) \cite{behrisch2011sumo} to accurately model the plant environment, vehicle behaviour, and fleet management system (FMS). By recreating the plant's environment and vehicle models in SUMO, the study was able to gather comprehensive data on both plant and vehicle states.

For effective traffic simulation in SUMO, a detailed map was created, including all vehicle points of interest such as potlines, cast houses, charging stations, and additional buildings for storage and maintenance. Each building was represented as a node in the network, with potlines marked by intersections representing clusters of pots. Charging stations were centrally placed on the map to avoid key areas of interest.

Vehicles followed simplified routines for task completion, halting near designated nodes with waiting times based on a Gaussian distribution proportional to task duration to enhance realism. Dijkstra's algorithm was chosen for vehicle routing due to its suitability for the network size and reliance on edge travel times, though this choice remains adaptable based on the specific system being simulated \cite{RoutingReview}.

Dijkstra's algorithm assesses the traverse time of each edge from the original location to determine the shortest path to the destination \cite{gupta2016applying}. By integrating the concept of visit values and a parameter ($W_{veh}$) indicating vehicle locations on the map, it becomes possible to identify a sub-optimal path that balances the shortest route with the least visited edges while avoiding other vehicles. Eq.\eqref{eq:dynamic_routing} illustrates how the traversal time ($T_{\Tilde{t}}$) of each edge dynamically updates at each simulation step, incorporating $W_{veh}$ if a vehicle occupies the edge, alongside the precise visit value calculated by the Forget Function ($FF(Edge_i|_{\Tilde{t}})$), weighted by $W{visit}$. These visit values are determined using the forget function described in eq.\eqref{eq:forget_function}.
\begin{align}
    \label{eq:dynamic_routing}
    &T_{\Tilde{t}} = T_t \cdot (1 +  W_{visit} \cdot FF(Edge_i|_{\Tilde{t}}) + W_{veh}) \\
    &FF(Edge_i|_{\Tilde{t}}) = 1+ \frac{FF(Edge_i|_{t})}{1 + e^{(K \cdot(\Tilde{t} - t - \Delta t))}}
    \label{eq:forget_function}
\end{align}

Here, $T_{\Tilde{t}}$ denotes the updated traverse time of the edge, while $T_t$ represents its original traverse time. Additionally, $W_{visit}$ serves as a weight determining the relevance of visit values for routing decisions, while $W_{veh}$ detects the presence of vehicles on the edge. The function $FF(Edge_i|_{\Tilde{t}})$ computes the exact visit value on the $i$-th edge at the current time, with $FF(Edge_i|_{t})$ representing the previous visit value at time $t$. Moreover, $K$ indicates the speed rate at which the visit value decreases over time, $\Tilde{t}$ stands for the current time step, $t$ signifies the last time step the edge was visited, and $\Delta t$ represents the time needed to halve the visit value.

Table \ref{tab:Veh_Plant_Params} reports the parameters employed during the simulation campaign.
\begin{table}[ht]
\centering
\caption{Vehicles Description and Plant Request Frequency. Courtesy of Techmo Car S.p.a.}
\label{tab:Veh_Plant_Params}
\begin{tabular}{|l|cccc|}
\hline
\multicolumn{5}{|c|}{Vehicle Parameters}\\
\hline
& APTV & FFV & MTV & UNIT \\
\hline
Acceleration & 0.8 & 0.8 & 0.8 & $[m/s^2]$\\
Deceleration & 4.5 & 4.5 & 4.5 & $[m/s^2]$\\
Max Speed & 22 & 15 & 15 & $[km/h]$\\
Vehicle Mass (Empty) & 17000 & 11000 & 22500 & $[Kg]$\\
Vehicle Mass (Full) & 31000 & 18000 & 52500 & $[Kg] $\\
Battery Capacity & 80 & 80 & 80& $[kWh]$\\
\hline
\multicolumn{5}{|c|}{Plant Request Frequency}\\
\hline
$AlF_{3}$ refill & \multicolumn{2}{c}{0.45 days} & \multicolumn{2}{c|}{10:57 [hh]}\\
Anode replacement & \multicolumn{2}{c}{0.0875 days} & \multicolumn{2}{c|}{2:06 [hh]}\\
Collect Aluminium & \multicolumn{2}{c}{0.24 days} & \multicolumn{2}{c|}{5:51 [hh]}\\
\hline
\end{tabular}
\end{table}

%% file: chapters/05_Results.tex
\section{Results}
\label{sec:Results}

\begin{figure}[ht]
    \centering
    \includegraphics[width=1\linewidth]{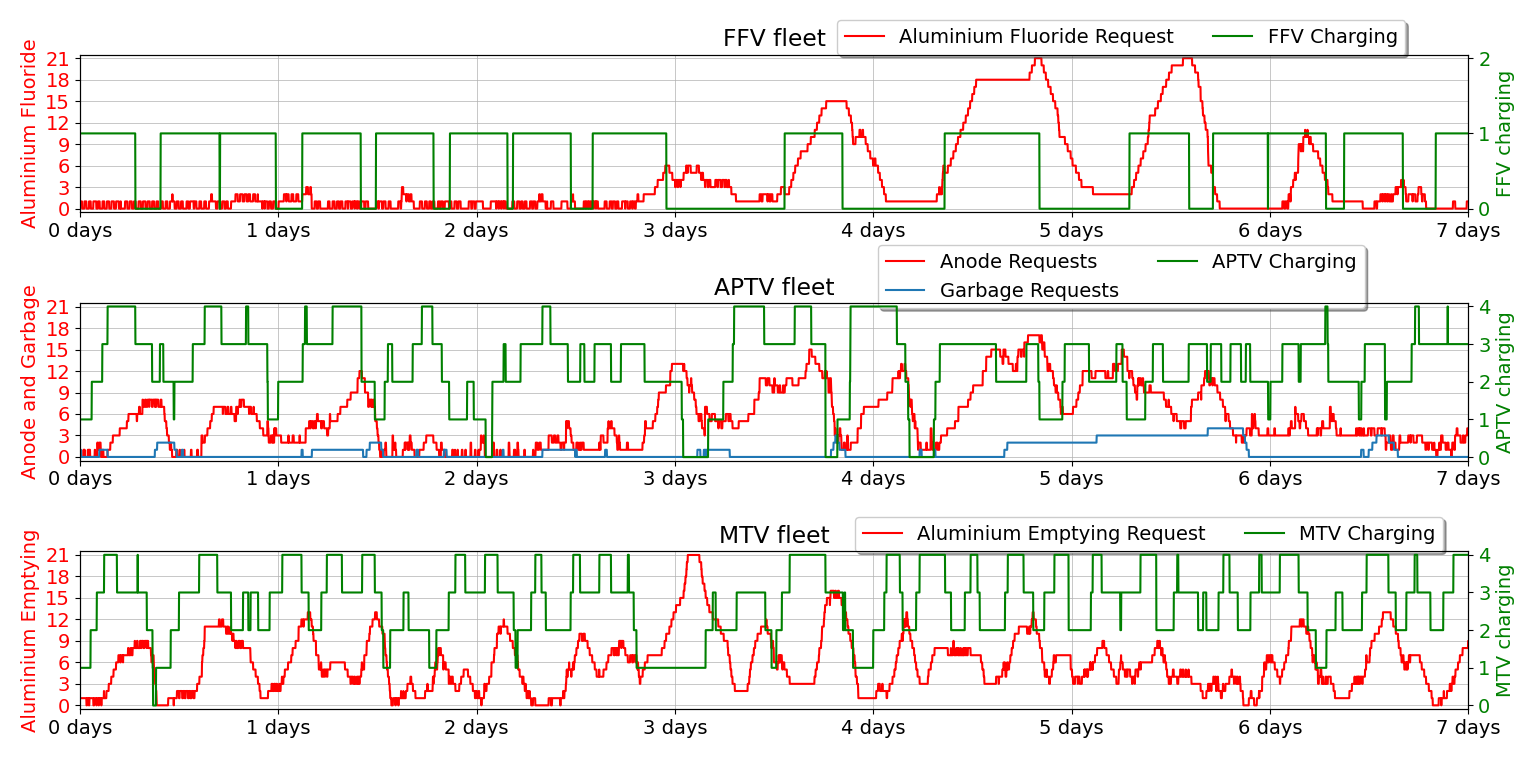}
    \caption{Comparison of plant requests and vehicle charging.}
    \label{fig:tasks_min}
\end{figure}

Testing various vehicle combinations, a suitable set has been identified that effectively prevents the accumulation of plant requests. Fig. \ref{fig:tasks_min} depicts the selected fleet of 2 \textit{FFV}, 4 \textit{APTV} and 4 \textit{MTV}, demonstrating its ability to fulfill all plant requests. The plots provide a comparison of active requests from the plant against vehicles charging their batteries throughout a one-week time frame.

The red line in each subplot represents the accumulation of the requests. The second subplot, associated with the \textit{APTV} fleet also provides the accumulation of the \textit{Garbage} requests in blue, while the green line indicates the vehicles that are undergoing a charging cycle.
The request fluctuations remain bounded throughout the entire simulation duration, preventing requests saturation. Only in rare instances do they reach their maximum value, typically due to an unfavorable convergence of demands from distant cells. 

\begin{figure}[ht]
    \centering
    \includegraphics[width=1\linewidth]{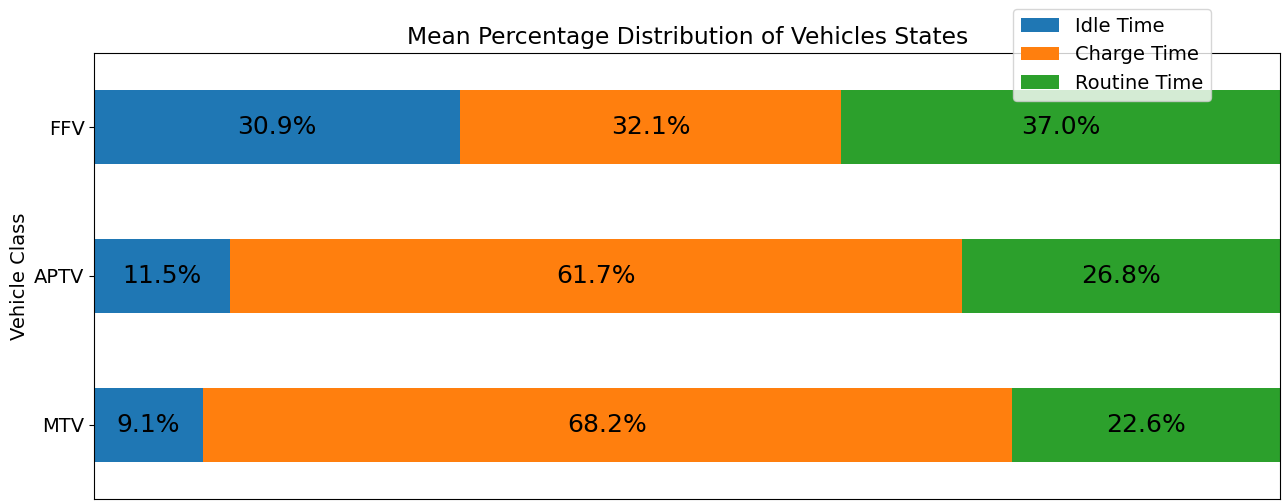}
    \caption{Mean Percentage Distribution of Vehicles States.}
    \label{fig:times_min}
\end{figure}

Collecting the amount of time each vehicle spends in the implemented states it has been possible to gather further insight on their behavior. Fig.\ref{fig:times_min} visualize the percentage of time the vehicles spend in each state averaged by the number of vehicles in the fleet.

It's interesting to note that the majority of vehicles spend over 60\% of their time charging their batteries. However, this trend doesn't apply to the \textit{FFV} fleet. This divergence can be attributed to the \textit{FFV} vehicles allocating more time to \textit{Idle} states, which also include the \textit{$Idle\_Charging$} state. 

\begin{figure}[ht]
    \centering
    \includegraphics[width=1\linewidth]{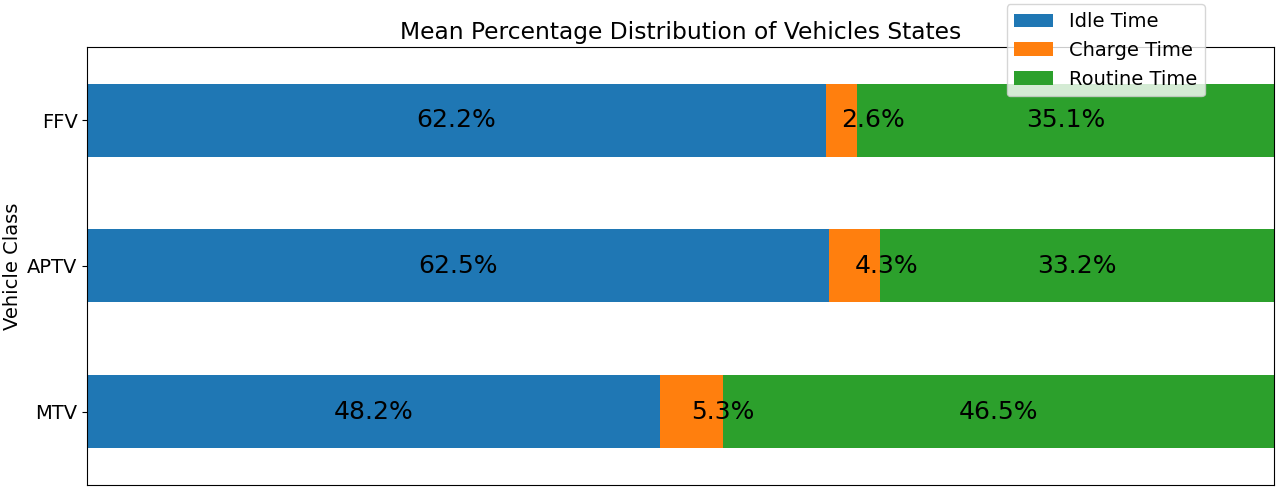}
    \caption{Mean Percentage Distribution of Vehicles States Replacing Batteries.}
    \label{fig:times_SBM}
\end{figure}

For this reason, it has been investigated a different scenario where the charging stations are equipped for battery swap, drastically cutting recharging times. 
Fig. \ref{fig:times_SBM} illustrates the mean, per vehicle class, of the percentage of time spent in each state when replacing the batteries. 
Investing in charging stations equipped with supplementary battery packs for fast replacement enables a significant reduction in vehicle downtime, ensuring quicker availability. As a result, the overall operational efficiency of the plant improves, potentially necessitating fewer vehicles to operate the facility effectively.

\begin{figure}[ht]
    \centering
    \includegraphics[width=1\linewidth]{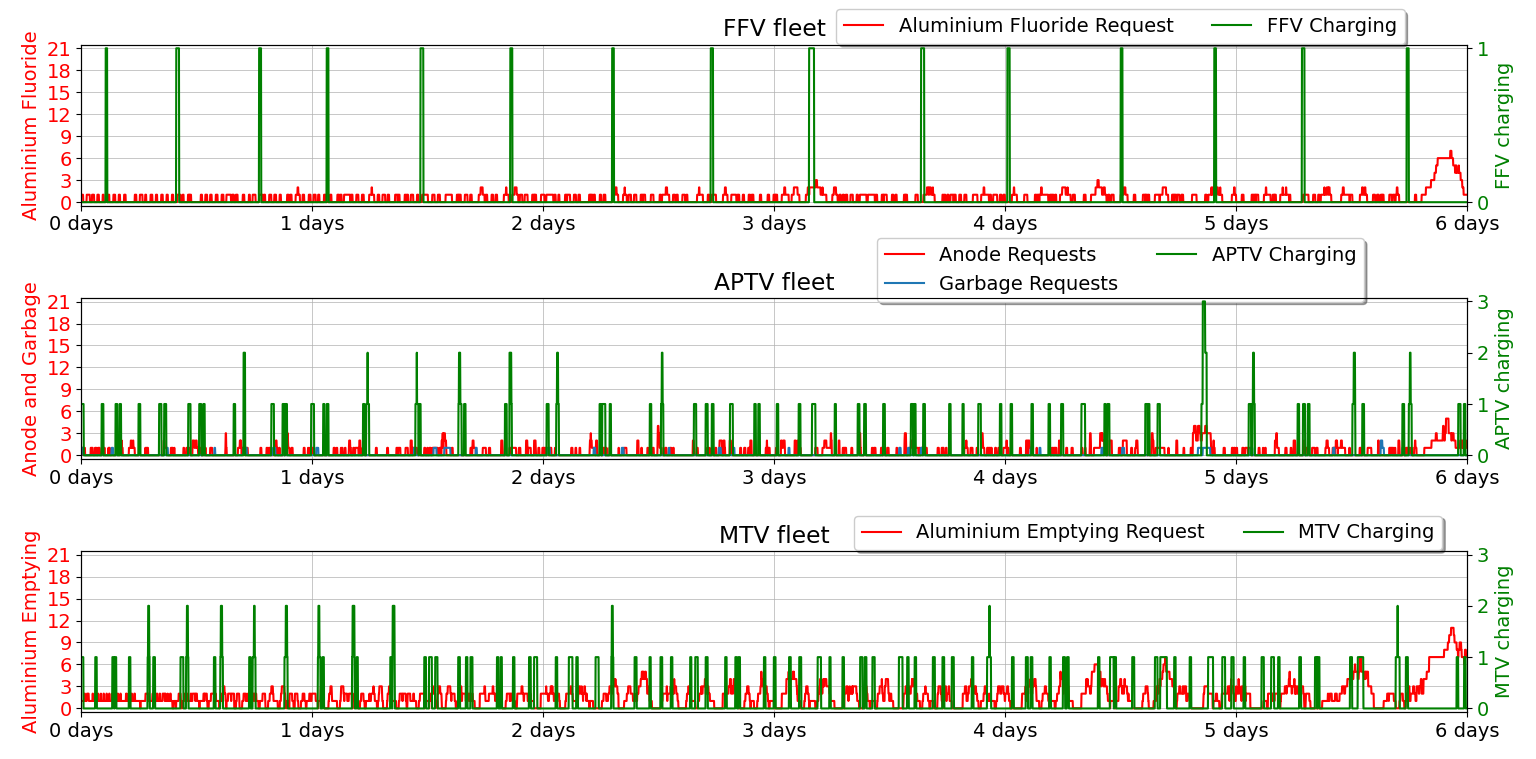}
    \caption{Comparison of plant requests and vehicle Replacing Batteries.}
    \label{fig:tasks_SBM}
\end{figure}

To validate the efficiency of replacing batteries, Fig. \ref{fig:tasks_SBM} shows the comparison of plant requests with vehicles replacing batteries. In this simulation, it has been employed an underestimated fleet, hence, a fleet with one vehicle less per class with respect to the simulation depicted in Fig. \ref{fig:tasks_min}.
With this approach, it becomes evident how vehicles can perfectly satisfy plant requirements, completing all tasks in a very short time and preventing requests accumulation even better than the minimum fleet estimated without battery swap.

By integrating the concept of "Visit Values" into vehicle routing using Dijkstra's algorithm, the study achieved a more balanced distribution of these values across the map, leading to increased street usage for navigation. Adjusting the weights $W_{surv}$ in the decentralized cost functions and $W_{visit}$ in the adapting traverse time allows for prioritizing either scouting under-visited streets or briefly charging the vehicle, and determining the significance of visit values in computing optimal paths, respectively.
Two simulations were conducted to illustrate the distribution of visit values. In the first simulation, routing based on visit values was disabled ($W_{visit} = 0$), and the surveillance task was deprioritized ($W_{surv} = 20$). In the second simulation, empirically determined values were used. Table \ref{tab:VisitValComparison} compares the percentage of edges with visit values below the threshold, where a lower percentage indicates better map coverage due to more edges being visited.

\begin{table}[h]
    \centering
    \caption{Comparison of the Visit Values of the edges at different steps.}
    \label{tab:VisitValComparison}
    \begin{tabular}{|l|cc|cc|cc|}
    \hline
        $W_{visit}$ & 0 & 0.1 & 0 & 0.1 & 0 & 0.1 \\
        $W_{surv}$  & 20 & 2 & 20 & 2 & 20 & 2 \\
        \hline
        Visit Value & \multicolumn{2}{c}{$\leq$ 30} & \multicolumn{2}{|c|}{$\leq$ 20}& \multicolumn{2}{c|}{$\leq$ 10}\\
        4 hours & 89.4\% & 95.3\% & 73.5\% & 72.3\% & 37\% & 20\% \\
        8 hours & 69.4\% & 79.4\% & 54.7\% & 54.1\% & 19.4\% & 10.6\% \\
        14 hours & 60\% & 38.2\% & 39.4\% & 13.5\% & 14.1\% & 1.2\% \\
    \hline
    \end{tabular}
\end{table}

%% file: chapters/06_Conclusions.tex
\section{Conclusion}
\label{sec:Conclusion}

This study introduces a versatile modular simulation tool designed for various contexts, providing comprehensive performance analysis from both vehicle and plant perspectives. Using an aluminum smelter with electric AGVs as a test case, initial trials identified the smallest fleet size required to prevent the accumulation of plant requests. However, a significant amount of time was spent by vehicles on charging.

To address this, battery-swapping was implemented, significantly reducing recharging time and enhancing efficiency. This improvement allowed for a smaller fleet size without sacrificing plant performance.

Additionally, a method based on Dijkstra's algorithm was developed to improve map coverage, optimizing routes by balancing the shortest paths with the least-visited streets and avoiding other vehicles. This method markedly increased street coverage, with only 1.2\% of streets visited fewer than 10 times after 14 hours, compared to 14.1\% without the method.

The study also highlights the tool's potential for further analyses, such as evaluating fleet robustness under vehicle failure scenarios. The use of cost functions within this architecture allows for task prioritization and tailored vehicle behavior by adjusting weight sets for different vehicle classes. While this flexibility enables extensive exploration of fleet and weight combinations, it also introduces complexity in finding optimal solutions, necessitating experience and time.